\documentclass{article}
\usepackage{spconf,amsmath,graphicx}
\usepackage{cite}
\usepackage{amsmath,amssymb}
\usepackage{amsfonts}
\usepackage[english]{babel}
\usepackage[autostyle]{csquotes}
\usepackage{efbox,graphicx}
\efboxsetup{linecolor=black,linewidth=.5pt}
\usepackage{amsbsy}
\usepackage{epstopdf}

\usepackage{multirow}
\usepackage{multicol}
\usepackage{subfigure}
\usepackage{textcomp}
\usepackage{stfloats}
\usepackage[export]{adjustbox}
\usepackage{epstopdf}
\usepackage{balance}
\usepackage{xcolor}
\usepackage{array}
\usepackage{longtable}
\usepackage{calc}
\usepackage{enumitem}
\usepackage{url}
\usepackage{amssymb}
\usepackage{gensymb}
\usepackage{subfig}
\usepackage{afterpage}
\usepackage{forest}
\usepackage{caption}
\usepackage{stackengine}
\usepackage{lipsum}
\usepackage{mathtools}
\usepackage[utf8]{inputenc}
\usepackage[T1]{fontenc}
\usepackage{cuted}
\usepackage{array,booktabs}
\newcolumntype{C}{>{$\displaystyle}c<{$}}
\usepackage{mathtools}
\newcolumntype{P}[1]{>{\centering\arraybackslash}p{#1}}
\def\L{{\cal L}}

\title{Transformer based self-context aware prediction for few-shot anomaly detection in videos}
\name{Gargi V. Pillai, Ashish Verma, Debashis Sen}
\address{ Department of E\&ECE, Indian Institute of Technology Kharagpur, India}
\begin{document}
%\ninept
%
\maketitle
\begin{abstract}
Anomaly detection in videos is a challenging task as anomalies in different videos are of different kinds. Therefore, a promising way to approach video anomaly detection is by learning the non-anomalous nature of the video at hand. To this end, we propose a one-class few-shot learning driven transformer based approach for anomaly detection in videos that is self-context aware. Features from the first few consecutive non-anomalous frames in a video are used to train the transformer in predicting the non-anomalous feature of the subsequent frame. This takes place under the attention of a self-context learned from the input features themselves. After the learning, given a few previous frames, the video-specific transformer is used to infer if a frame is anomalous or not by comparing the feature predicted by it with the actual. The effectiveness of the proposed method with respect to the state-of-the-art is demonstrated through qualitative and quantitative results on different standard datasets. We also study the positive effect of the self-context used in our approach. 

%In this paper, we are proposing a transformer-based video anomaly detection model where the training of the model is done only for normality class in single video with first few non-anomalous consecutive frames, hence \textit{few-shot}, for frame-level anomaly detection on that video. This proposed few-shot model eliminates the need for a large training dataset. 
%The results are compared with the state-of-the-art methods using AUC and ROC curve. Video anomaly detection plays an essential role in public security. Recently, majority of video anomaly detection approaches are based on deep learning networks that demand a huge amount of training data. 
% In this paper, we are proposing a transformer based unary few shot video anomaly detection based on the next video frame prediction. We combine spatial features extracted using pretrained ResNet and optical flow based temporal features to represent video frames. Most of the deep learning networks do training on huge datasets. We are training the transformer with a few consecutive non-anomalous frame features to predict the next frame features to be used in anomaly detection for each video, thus reducing the size of training dataset significantly. The effectiveness of the proposed method is demonstrated by generating qualitative and quantitative results on different standard datasets. The results are compared with the state-of-the-art methods using AUC and ROC curve.
\end{abstract}
\section{Introduction}
\label{sec:intro}
\vspace{-.2cm}
%difficult to manually analyze and identify anomalies from a large amount of video surveillance data. So it is necessary to
With an increase in demand of video surveillance systems for video anomaly detection (VAD), it is becoming increasingly important to develop intelligent surveillance systems to automatically detect anomalies in different kinds of scenarios~\cite{wang2021robust}. Overcoming the challenging nature of VAD, many recent approaches on VAD based on deep learning have achieved significant improvements over classical approaches \cite{zhu2018real, ganokratanaa2020unsupervised}. These approaches can be generally categorized as reconstruction based \cite{zhu2018real, nawaratne2019spatiotemporal} and prediction based approaches \cite{wang2021robust, ganokratanaa2020unsupervised}. Most deep learning based VAD approaches model normality by training on data without anomaly and infers abnormality in the testing data using prediction or reconstruction error.

Reconstruction based VAD approaches detect anomalies by reconstruction of video frames, where low and high reconstruction errors represent normality and abnormality, respectively. Reconstruction based deep approaches largely include those based on autoencoders (AE)\cite{zhu2018real, chang2022video} and its variants such as convolutional autoencoders (CAE) \cite{li2020spatial, wang2021intermediate, asad2021anomaly3d, hao2022spatiotemporal}. Prediction based deep approaches detect anomalies by predicting current frame features using that of previous frames. A low /high prediction error signifies the presence of normal /abnormal events. Prediction based approaches mostly include those using autoregressive models (AR) \cite{abati2019latent, pillai2021anomaly}, convolutional long short-term memory (CLSTM) \cite{luo2017remembering} and generative adversarial networks (GAN) \cite{liu2018future, chen2020anomaly, ganokratanaa2020unsupervised, zhang2020normality}. Prediction-based approaches have been found to be successful in learning the invariances related to the temporal changes when an anomaly is not present, and hence, they perform well in VAD~\cite{cai2021video}.
%Due to exceptional generative capability of deep neural networks (DNN), the reconstruction ability of the approaches also increase which reduces the discriminative power of such reconstruction based approaches. This can be solved to some extent by prediction-based approaches by forcing the model to strengthen the time connection during next frame prediction improving the detection effect \cite{cai2021video}.

\begin{figure}[t]
\centering
%\captionsetup{justification=centering}
\setlength{\fboxsep}{0.008pt}
\setlength{\fboxrule}{0.8pt}\includegraphics[scale=.9]{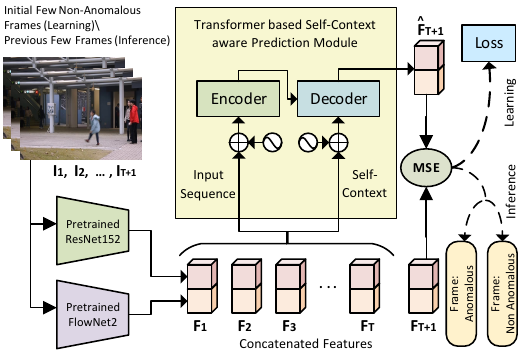}
	\vspace{-0.3cm}
\caption{\small{Schematic of the proposed self-context aware video anomaly detection through one-class few-shot learning. $ F_1, F_2, \cdots, F_{T+1}$ represent the concatenated spatial and temporal features of a set of $T+1$ consecutive frames $I_1, I_2, \cdots, I_{T+1}$ in the video. Previous $T$ frames are used for anomaly detection in the current $(T+1)^\textrm{th}$ frame, by comparing its predicted feature vector to the actual during inference. The same loss is used in the network learning using its initial few non-anomalous frames for the video at hand.}}
\label{bd}
\vspace{-0.6cm}
\end{figure}
%where spatial and temporal features of frames $I_0, I_2, \cdots, I_{t-1}$ are concatenated to obtain feature vectors $F_0, F_1, \cdots, F_{t-1}$ on which next frame prediction module is applied to obtain predicted feature vector $\hat{F_t}$.
%most of the deep learning based works mentioned above perform well for VAD. However, these methods require a large dataset for training.
\par Most of the above deep learning based approaches that perform well in VAD are trained on a dataset of videos. Given the varying kinds of anomaly from one video to another, the scope of designing a model that can be trained through few-shot learning in the single video at hand would be interesting to explore. This can significantly reduce the training data requirement and be suitable for videos in different scenarios. In the last few years, transformers have been found to be extremely effective in sequence prediction~\cite{wang2021end}. In transformers, all sequence positions relevant to the encoder's input are attended by every sequence position related to the data fed into the decoder~\cite{vaswani2017}. This aspect can be leveraged to capture both the relation among and the context of the non-anomalous frames of a video through only few-shot learning, as transformers are highly effective in modeling dependencies within samples in a sequence.

In this paper, we propose an approach for anomaly detection in a video based on one-class few-shot learning of the transformer network only using the initial few frames in that video. Our transformer network's encoder gets the features of a few consecutive video frames as the input and its decoder predicts the feature vector of the subsequent frame in the video as the output. The input features into the encoder are employed as the input to the decoder as well, which allows them to act as a self-context attending over all the frames in the input sequence. 

The one-class few-shot learning is performed considering the initial few non-anomalous video frames for the encoder and decoder inputs. The transformer learns the relation among the non-anomalous frames to predict the feature of the subsequent non-anomalous video frame, given the self-context. After the learning using the few initial non-anomalous frames of the video at hand, anomaly detection is performed in the rest of the video or continuously for the duration required. A frame is marked as anomalous when its feature vector predicted by our network using a few previous frames differs from the actual feature vector. During this inference, we use the actual or predicted (if the actual is anomalous) features of the previous few frames.

The feature vector of a video frame is obtained by concatenating the spatial and temporal features extracted using the pre-trained ResNet152 \cite{he2016deep} and FlowNet2 \cite{ilg2017flownet} networks, respectively. Additionally, temporal consistency is also imposed to reduce false positives during the anomaly detection.
The main contributions of this paper are: 
\begin{itemize}[noitemsep]
\vspace{-.2cm}
    \item Video-specific one-class few-shot learning based VAD, where the non-anomalous nature is learned for the video at hand without any training on a dataset.
    \item The use of the transformer network for prediction-based VAD in a way where its sequence dependency modeling capabilities are thoroughly exploited under the attention of a self-context.
    %It makes use of the sequence modelling ability of transformer network for video frame prediction. 
    %\item To the best of our knowledge, this is the first work that employs a few-shot learning in VAD, which reduces the need of a huge dataset that is usually required in DNN-based VAD.
\end{itemize}
\vspace{-.2cm}
We demonstrate the superiority of our approach over the state-of-the-art through experimental results on standard datasets with different kinds of videos. An ablation study reveals that the use of the proposed self-context provides a significant boost to our VAD performance.

Section~\ref{sec:approach} describes our approach, the experimental results are given in Section~\ref{sec:result}, and Section~\ref{sec:conclusion} concludes the paper.
\section{The Proposed Self-Context Aware Prediction for Video Anomaly Detection}
\label{sec:approach}
A pictorial overview of our proposed approach is shown in Fig~\ref{bd}. %{\color{blue} The proposed method consists of  feature extraction and predictive modelling of the extracted features for anomaly detection.} 
The central issue of the VAD problem in our hand can be formulated as: Given a few successive previous video frames represented by their features, we need to estimate the features in the next frame to decide if it is anomalous or not. %non-anomalous
%a video clip with features of a few successive non-anomalous video frames, we need to predict feature of future frame to be compared with the actual frame feature for VAD.
%\par Usually, normal frames have similar features as its neighbourhood. In this work, we are assuming that a frame is anomalous when it has significantly different characteristics from normal events. Therefore, frame are classified as anomaly. Temporal consistency is considered by considering only those frames anomalous if neighbourhood frames are also anomalous.
\subsection{Video Frame Feature Extraction}
\label{sec:Feature Extraction}
We denote the feature vector representing the $t^\textrm{th}$ frame in a sequence of video frames as $F_t$, which is obtained by concatenating spatial and temporal features following usual norms~\cite{li2020spatial,doshi2020continual}. We consider the features $R_{i}, i=1,2,\ldots,512,$ extracted using the pretrained ResNet512 of \cite{he2016deep} on the video frame as the spatial features, and the features $O_{i}, i=1,2,\ldots,mn,$ extracted using pretrained FlowNet2 of \cite{ilg2017flownet} on the video with frame size $m \times n$.
%The definition of anomaly is broad; hence it is not ideal to consider only temporal or spatial information of the videos. In our proposed model, we extract both temporal and spatial features for video frame representation on which transformer-based prediction model is employed for anomaly detection. The spatial features $R_{1}, R_{2}, ...,R_{512}$ extracted from pretrained ResNet and optical flow based temporal features $O_{1}, O_{2}, ..., O_{mn}$ extracted using pretrained FlowNet2 are used to represent the video frames of size $m \times n$. These features are concatenated to obtain final feature vector $F_1$, $F_{2}$, ..., $F_{T}$ as shown in Fig~\ref{bd}. 
  \begin{figure}[t]
\centering
\setlength{\fboxsep}{0.008pt}%
		\setlength{\fboxrule}{0.8pt}\includegraphics[scale=.9]{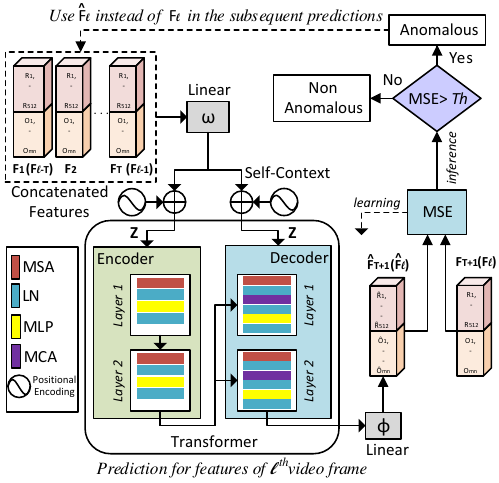}
	\vspace{-0.3cm}
\caption{ \small{Architecture of the proposed transformer based prediction module that leverages a self-context.}}
\label{pred}
	\vspace{-0.6cm}
\end{figure}
\subsection{Our Transformer based Prediction for VAD}
The architecture of the transformer based self-context aware prediction module is shown in Fig.~\ref{pred}, on which a one-class few-shot learning strategy is applied for anomaly detection during the inference.
We consider a transformer~\cite{vaswani2017} with substantially fewer number of encoder and decoder layers than usual, which we find is sufficient for our learning problem. While the encoder of the network contains the two standard modules of Multi-head Self-Attention (MSA) \& Layer Normalization (LN), and Multi-Layer Perceptron (MLP) \& LN both repeated only twice, the decoder contains the aforesaid modules along with the standard module of Multi-head Cross-Attention (MCA) \& LN again all of them repeated only twice (See Fig~\ref{pred}). 
%As videos can be represented as a series of frames indexed in time, the sequence modeling capability of transformer is used for learning the dependencies among video frames using frame feature vectors. The transformer model follows an encoder-decoder architecture. Each encoder-decoder layer consists of $h$ heads of Multi-head Self-Attention (MSA), Layer Normalization (LN), and Multi-Layer Perceptron (MLP). Transformer decoder consists of an additional layer i.e. $h$ heads of Masked Multi-head Self-Attention (MMSA). 

The feature vectors extracted from a few consecutive video frames are given as the sequence input to the encoder, acting upon which the decoder only predicts the feature vector of the next video frame after the sequence. The same input sequence is also fed into the decoder, and therefore, a learned representation of the input sequence (from encoder) is attended by another learned representation of the same sequence (from decoder) at the MCA modules in the decoder forming the self-context.

Consider a sequence of feature vectors extracted from $T$ consecutive video frames as $F_1, F_2, \ldots, F_{T}$, and we estimate $\hat{F}_{T+1}$ in an attempt to predict $F_{T+1}$ from $F_t, t=1, 2, \ldots, T$. The feature vectors $F_1, F_2, \ldots, F_{T}$ are subjected to a learnable linear layer $\omega$ obtaining $T$ vectors of a dimension as required by the transformer. Positional encoding \cite{vaswani2017} is then applied to embed the sequence position information yielding $T$ position-aware feature vectors $z_1, z_2, ..., z_{T}$, which are fed as the sequence input to the transformer's encoder $\Gamma_{E}$ consisting of just $2$ layers of the standard MSA \& LN and MLP \& LN modules of \cite{vaswani2017}. The encoder provides a latent representation $u$ of the input feature vector sequence $F_t, t=1, 2, \ldots, T$. Therefore, we have:
\begin{eqnarray}
    z_t&=&\omega\left(F_{t} \right) + P(t), \ \ \ \ t \in 1, ..., T\\ 
    u&=&\Gamma_{E}(\mathbf{Z}),  \ \ \ \ \mathbf{Z}=\{ z_1, z_2, ..., z_{T} \}
\end{eqnarray}
where $P(t)$ denotes the positional code value for the $t^{th}$ feature vector $F_t$. The output $u$ from the last layer of the encoder is fed into the decoder at all its MCA modules.
%The position-aware sequence of feature vectors, $\mathscript{Z}=\{ z_1, z_2, ..., z_{T} \}$, is then passed through a transformer encoder $\Gamma_{{E}}$ consisting of $L_{E}$ layers which results in an abstract representation $\mathscript{U}$ of the input sequence $F_1, F_2, ..., F_{T}$. The process of generation of the $\mathscript{U}$ through transformer encoder $\Gamma_{{E}}$ is as follows:
%\begin{align}
 %  & \vartheta_{t} = \omega\left(F_{t} \right) \ \ \  \ \ \ \forall t \in 1, ..., T \\
  % & z_t=\vartheta_{t} + PE_{t} \ \ \ \ \forall t \in 1, ..., T \\
  % &\mathscript{U} = \Gamma_{E}(\mathscript{Z}),  \ \ \ \ \mathscript{Z}=\{ z_1, z_2, ..., z_{T} \}
%\end{align}
%\begin{equation}
%    \vartheta_{t} = \omega\left(F_{t} \right) \ \ \  \ \ \ \forall t \in 1, ..., T
%\end{equation}
%\begin{equation}
%    z_t=\vartheta_{t} + PE_{t} \ \ \ \ \forall t \in 1, ..., T 
%\end{equation}
%\begin{equation}
%    \mathscript{U} = \Gamma_{E}(\mathscript{Z}),  \ \ \ \ \mathscript{Z}=\{ z_1, z_2, ..., z_{T} \}
%\end{equation}
%The sequence of feature vectors $F_1, F_2, \ldots, F_{T}$ subjected to the learnable linear layer $\omega$ for dimension adjustment as required by the decoder and then embedded with positional encoding to obtain $T$ position-aware feature vectors $y_1, y_2, ..., y_{T}$, which 

The position-aware feature vectors $z_1, z_2, ..., z_{T}$ are also fed to the transformer's decoder $\Gamma_{D}$. The decoder's output is subjected to a learnable linear layer $\phi$, which provides only the estimate $\hat{F}_{T+1}$ for the input feature vector sequence $F_t, t=1, 2, \ldots, T$. Therefore, we have:
\begin{eqnarray}
    \hat{F}_{T+1}&=&\phi(\Gamma_{D}(\mathbf{Z},u)),  \ \ \ \ \mathbf{Z}=\{ z_1, z_2, ..., z_{T} \}
\end{eqnarray} 
Note that, as our transformer predicts only $\hat{F}_{T+1}$ from the input sequence $F_t, t=1, 2, \ldots, T$, which does not contain $F_{T+1}$, we do not require the mask function~\cite{vaswani2017} used in the standard transformer. Further, we also do not require any recursion, where the output of the decoder is supplied as its input. The decoder $\Gamma_{D}$ consists of just 2 layers of the standard MSA \& LN, MCA\& LN, MLP \& LN modules and both the MCA modules receive $u$ as the `value' and `key' quantities from the encoder $\Gamma_{E}$ to be attended by the `query' from $\Gamma_{D}$ itself.

\textbf{\textit{One-class Few-shot Learning:}} 
For the video at hand, we consider that an initial consecutive sequence of $N$ frames (few-shot) are non-anomalous (one class), and their feature vectors $F_1, F_2, ..., F_{N}$ are used to pool multiple sets of $T+1$ consecutive feature vectors for the transformer learning. A set of consecutive $T$ video frames' feature vectors are used to predict the ${(T+1)}^{th}$ frame's feature vector. In an epoch, every set of $T+1$ consecutive feature vectors available from the $N$ feature vectors are considered in random order with one set representing a learning iteration. For the learning, we consider the mean square error (MSE) loss function ($\sim$ norm of vector difference) between a predicted feature vector $\hat{F}_{T+1}$ and the corresponding actual feature vector ${F}_{T+1}$ as follows:
\begin{equation}
  \small  L_{MSE} = \frac{1}{D}\sum_{j=1}^{D}(F_{T+1}(j)-\hat{F}_{T+1}(j))^2 
\end{equation} 
where, $F_{T+1}$ is of dimension $D = 512 + mn$.
%Following the approach discussed above, the proposed model gives predicted sequence of feature vectors $\hat{F}$. We are dividing each video into overlapping video clips having $T$ frames with stride $1$ to predict ${(T+1)}^{th}$ frame feature vector. This ensures that the predicted $\hat{F}_{T+1}$ for $k^{th}$ and $({k+1})^{th}$ iteration are representing the $t^{th}$ and $({t+1})^{th}$ frames of the video. 
%Hence, our objective is to reduce the mean square loss (MSE) between predicted feature vector $\hat{F}_{T+1}$ and actual feature vector ${F}_{T+1}$ of ${(T+1)}^{th}$ frame. The loss $L_{MSE}$ for training of the proposed model is calculated as:
 
\textbf{\textit{Inference for Anomaly Detection:}} 
All the frames in the video except the few initial ones used for learning are considered here for anomaly detection. To detect whether a frame is anomalous or not, the previous $T$ frames are considered for the input to the transformer. As our transformer is trained to predict the non-anomalous feature vector of the current frame given a sequence of previous frames as input, the current frame will naturally be marked as anomalous if the predicted feature vector differs from its actual feature vector. We compute the difference (anomaly score) as the $L_{MSE}$ (norm square of vector difference) between the actual and the predicted feature vectors. If for the $t^\textrm{th}$ frame, $L_{MSE}(t)\geq Th$, then the frame is marked as an anomaly, where $Th$ is the post-convergence average $L_{MSE}$ for all the $N-T$ predictions performed during the learning in the video at hand. 
%To infer whether a frame is an anomaly or not we consider mean square error $L_{MSE}$ as anomaly score for that frame. The frame is classified as anomaly if the condition $L_{MSE} \geq Th$ is satisfied. $Th$ is the threshold calculated by considering the average of loss $L_{MSE}$ for initial non-anomalous $N$ consecutive frames. 
Note that, as our transformer works with non-anomalous features of consecutive frames as inputs for the prediction, the predicted feature vectors of the frames already marked as anomalous are considered in subsequent predictions for anomaly detection in the forthcoming frames, instead of the corresponding actual feature vectors. Finally, a temporal consistency is imposed by considering a frame as anomalous only when the frames in their immediate temporal neighborhood are also anomalous.
%Further, to make sure that only previous non-anomalous consecutive frames are used for next frame prediction, the feature vector of frames that are classified as anomaly is replaced by the predicted feature vector.
\begin{table}[htb!]
	\begin{center}
	\vspace{-.2cm}
	\caption{\label{table} \small{Result comparison for UCSD Ped2, CUHK Avenue, and ShanghaiTech Campus datasets using Frame-level AUC ($\%$)}}
	\vspace{-.2cm}
	\setlength\arraycolsep{3pt} 
	\begin{tabular}{| m{3.5cm}| P{.5cm} P{.8cm} P{2.1cm}|}
 \hline
 \small{Method} &  \small{Ped2} & \small{ Avenue} &  \small{ShanghaiTech}\\\hline
 \small{ConvLSTM-AE \cite{luo2017remembering}, 2017}  & \small{88.1}&\small{77.0}&\small{$\times$}\\
\small{Zhu et al. \cite{zhu2018real}, 2018} & \small{97.1}&\small{$\times$}&\small{$\times$}\\
\small{FFP+MC \cite{liu2018future}, 2018}& \small{95.4} &\small{85.1} &\small{72.8	}\\
\small{ISTL \cite{nawaratne2019spatiotemporal}, 2019} & \small{91.1}&\small76.8&\small{$\times$}	\\
\small{Abati et al. \cite{abati2019latent}, 2019}& \small{95.4}&\small{$\times$}&\small{72.5}\\
\small{Chen et. al. \cite{chen2020anomaly}, 2020} & \small{96.6} &\small{$\times$} &\small{$\times$}\\
\small{DSTN \cite{ganokratanaa2020unsupervised}, 2020} &  \small{95.5}& \small{87.9}&\small{$\times$}	\\ 
\small{Multispace \cite{zhang2020normality}, 2020}  & \small{95.4}&\small{86.8}&\small{73.6}\\
\small{ST-CaAE \cite{li2020spatial}, 2020} & \small{92.9}&\small{83.5}&\small{$\times$}\\
\small{Doshi et al. \cite{doshi2020continual}, 2020} &\small{97.8}&\small{86.4}&\small{71.6}	\\
\small{Wang et al. \cite{wang2021intermediate}, 2021}& \small{96.0}&\small{86.3}&\small{74.5}\\
\small{Anomaly3D \cite{asad2021anomaly3d}, 2021}& \small{95.8}&\small{89.2}&\small{\textbf{80.6}}\\
\small{Msm-net \cite{cai2021video}, 2021} & \small{96.8}&\small{87.4}&\small{74.2}\\
\small{ROADMAP \cite{wang2021robust}, 2021} & \small{96.3}&\small{88.3}&\small{76.6}	\\ 
\small{TRD \cite{pillai2021anomaly}, 2021}  & \small{98.2}&\small{89.3}&\small{80.2}
\\
\small{Chang et al. \cite{chang2022video}, 2022} & \small{96.7}&\small{87.1}&\small{73.7}	\\ 
\small{STCEN \cite{hao2022spatiotemporal}, 2022}& \small{96.9}&\small{86.6}&\small{73.8}\\
 \hline
\small{\textbf{Ours}} & \small{\textbf{98.6}}&\small{\textbf{89.4}}&\small{\textbf{80.6}}	\\  \hline
	\end{tabular}
\end{center}
\vspace{-.6cm}
\end{table}
\section{Experimental results}
\label{sec:result}
\textbf{Datasets and Evaluation Metrics: }We present the qualitative and quantitative results of our VAD approach considering videos from three frequently used datasets, namely, the UCSD Pedestrian 2 (Ped2) dataset \cite{ucsd}, the CUHK Avenue dataset \cite{lu2013abnormal}, and the ShanghaiTech Campus dataset \cite{liu2018future}. The anomaly in the Ped2 dataset is non-pedestrians in a pedestrian path,  anomalies in the Avenue dataset are running, moving in wrong direction and strange actions such as throwing, dancing etc., and anomalies in the ShanghaiTech dataset include running, skating and biking. We compare our model with several state-of-the-art methods using AUC at frame level \cite{zhang2020normality, abati2019latent} and ROC curve, which are standard measures \cite{zhang2020normality, abati2019latent}.
\par \textbf{Implementation Details:} In the proposed network, we consider just $2$ layers of the modules in the transformer's encoder and decoder with the input dimension as $512$. The number of heads used in the various multi-head attention modules is $2$. We take $N$ = 50, which is the number of initial non-anomalous frames considered for learning, and consider $T$ = 10, which is the number of previous frames given at the input to perform the prediction related to the current frame. We use the Adam optimizer (lr = 0.01, $\beta$ = (0.9,0.98)) for the training, which runs for $100$ epochs in a video.
 %We train our model with $N$ = 50 non-anomalous consecutive frames for each video used for VAD. We are dividing each video into overlapping video clips having $T$ = 10 frames with stride $1$ to predict ${(T+1)}^{th}$ frame.

 \begin{figure*}[t]
	 \vspace{-.4cm}
		\centering
		\begin{tabular}{p{4cm} p{4cm} p{4cm} p{4cm}}
\efbox{\includegraphics[width=4.1 cm, height=2.7 cm]{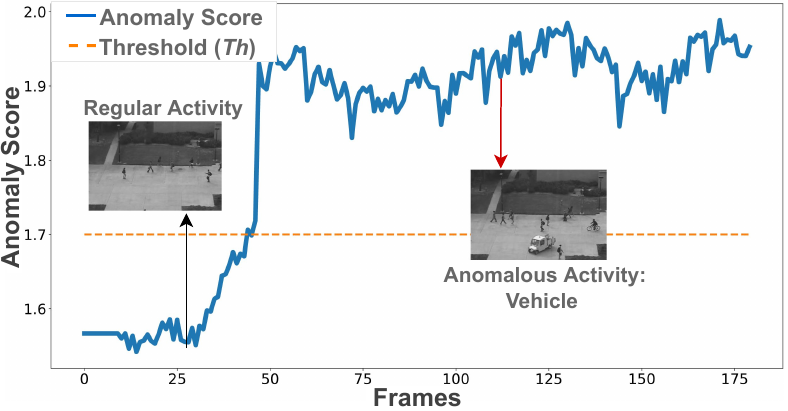}}
%&\efbox{\includegraphics[width=3.7 cm, height=2.2 cm]{fig/ucsd_12_result_modified.pdf}}
&
\efbox{\includegraphics[width=4.1 cm, height=2.7 cm]{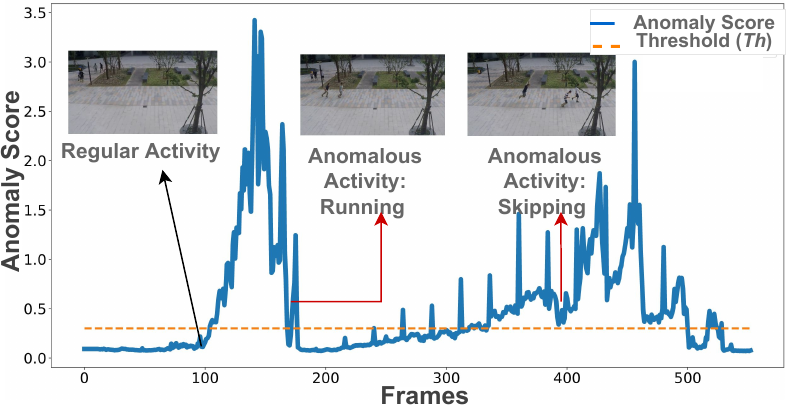}}
%&\efbox{\includegraphics[width=3.7 cm, height=2.2 cm]{fig/avenue_20_result_modified.pdf}}
&
\efbox{\includegraphics[width=4.1 cm, height=2.7 cm]{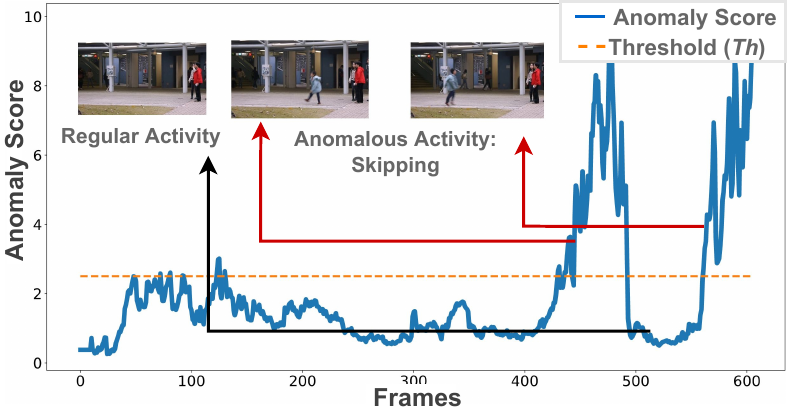}}
%&\efbox{\includegraphics[width=3.7 cm, height=2.2 cm]{fig/shanghaitech_08_0044_result_modified.pdf}}
&
\efbox{\includegraphics[width=4.1 cm, height=2.7 cm]{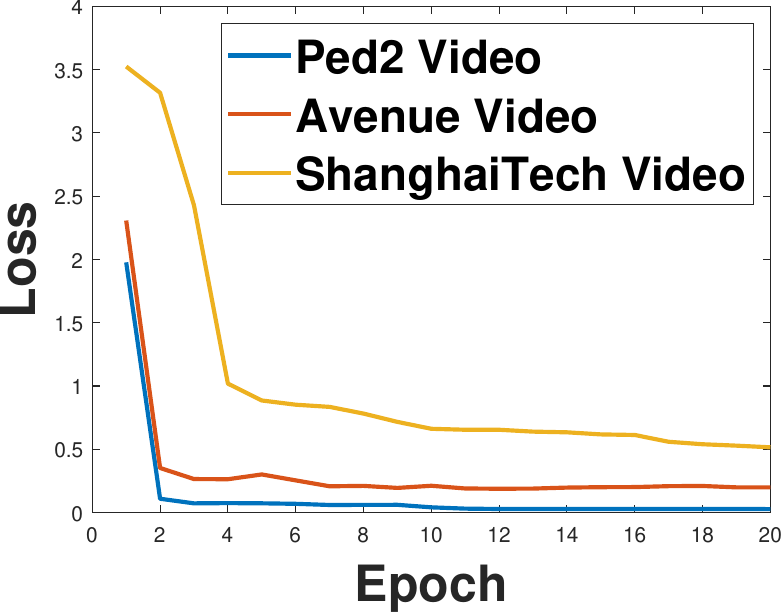}}\\
\hspace{2cm}{\small{(a)}}&\hspace{2cm}{\small{(b)}}&\hspace{2cm}{\small{(c)}}&\hspace{2cm}{\small{(d)}}
		\end{tabular}
		\vspace{-.4cm}
		\caption{\small{Anomaly scores of few frames in videos from (a) UCSD Ped2 dataset, (b) ShanghaiTech Campus dataset, and (c) CUHK Avenue dataset. (d) Learning curves for a video each from the 3 datasets.}}
				\label{loss}		\vspace{-.6cm}
\end{figure*}
 \begin{figure}[h]
\vspace{-.2cm}
		\centering
		\setlength\tabcolsep{1pt}
		\begin{tabular}{ c c c}			\includegraphics[width=2.8cm,height=3cm]{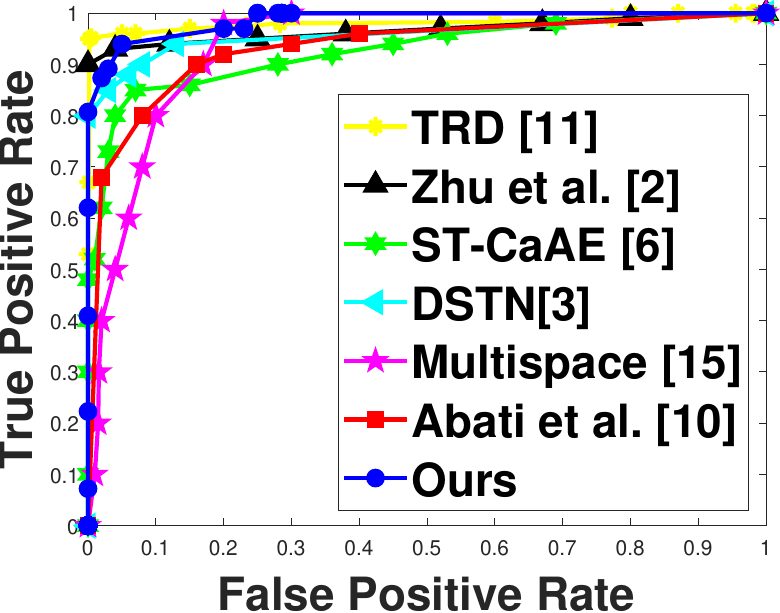} &
		\includegraphics[width=2.8cm,height=3cm]{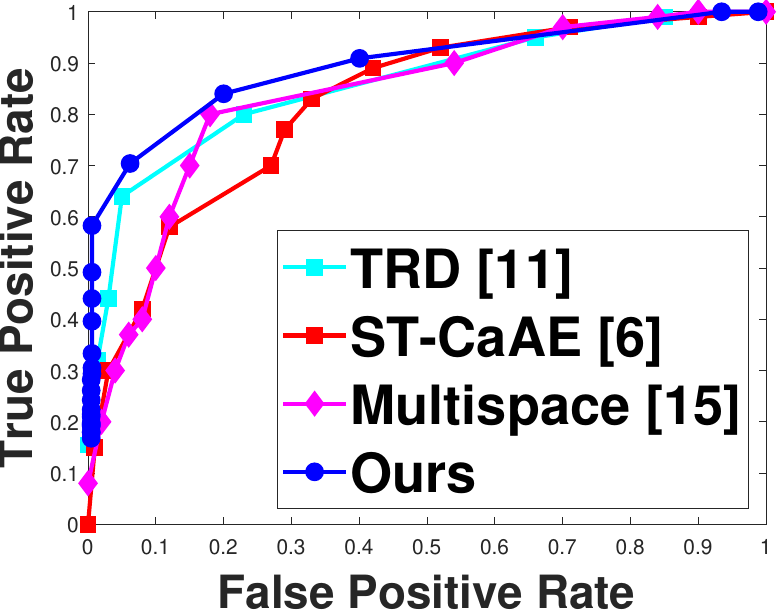} &
		\includegraphics[width=2.8cm,height=3cm]{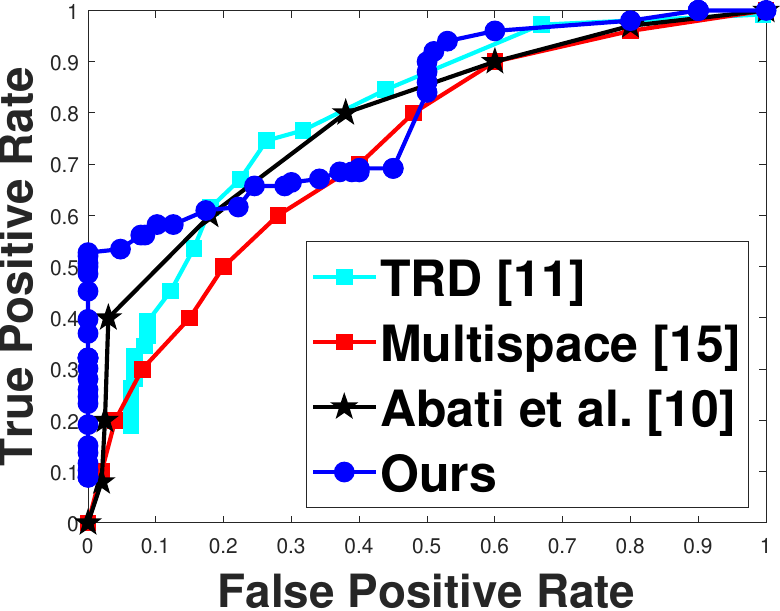}
			\\
		\small{(a)}&\small{(b)}&\small{(c)}\\
		\end{tabular}
		\vspace{-.4cm}
		\caption{\small{Frame-level ROC curves for (a) UCSD Ped2, (b) CUHK Avenue, and (c) ShanghaiTech Campus datasets.}}
		\label{result-roc}
		\vspace{-.6cm}
	\end{figure}
\subsection{Results and Discussion} 
Our approach is quantitatively compared with around $17$ existing approaches including the state-of-the-art on the three datasets. Table~\ref{table} shows the quantitative results of the different approaches including ours. We observe that our approach performs the best in terms of frame-level AUC on the Ped2 and Avenue datasets, and performs at par with the state-of-the-art on the ShanghaiTech dataset. The ROC curves for the three datasets shown in Figs.~\ref{result-roc}(a), (b) and (c) correspond to some of the AUC results in Table~\ref{table} including ours, where our approach is found to perform at least as good as the other compared methods. From the above observations, we find that the performance of our approach surpasses the state-of-the-art on the three datasets, which contain varied types of normality ranging from low to high complexity. For qualitative analysis, we visualize the anomaly score ($L_{MSE}$) for a set of frames in a video each from the three datasets in Figs.~\ref{loss}(a), (b) and (c), where the corresponding thresholds on the anomaly score to detect anomalous activities are also given. As observed in the figure, our approach successfully detects a vehicle in pedestrian path, skipping and running as anomalous events as required.
%As observed in the Fig ~\ref{loss} (a), various non-pedestrian events such as driving, skating were detected on pedestrain path in UCSD-Ped2 dataset. Qualitative results for CUHK Avenue dataset shows that anomalous activities such as dancing, throwing were detected successfully as seen in Fig ~\ref{loss} (b). Anomalous activities such as skating, running, skipping were detected in ShanghaiTech Campus dataset as shown in Fig ~\ref{loss} (c).
%with varying frames on UCSD-Ped2, CUHK Avenue, and ShanghaiTech Campus datasets in Figs~\ref{loss} (a), ~\ref{loss} (b), and ~\ref{loss} (c), respectively.  A frame is classified as an anomaly or not by thresholding on the anomaly score $L_{MSE}$.
In Fig.~\ref{loss}(d), loss curves of our one-class few-shot learning on videos from the three datasets are given, which shows its efficiency represented by the fast convergence.
%The effectiveness of our model OCFS-TransVAD is observed from the fast converging learning curve shown in Fig ~\ref{learning}, which implies that our model is efficiently learning the representation of normality in a video from training on few consecutive non-anomalous frames. 

% as shown by both the quantitative and qualitative results.It is observed that the AUC ($\%$) value for UCSD-Ped2 dataset is higher compared to other datasets which is due to the low level complexity for modelling the normality in the dataset. For example, the normality in UCSD-Ped2 dataset is pedestrians walking in same background with non-changing foreground size, while normality in CUHK Avenue dataset has objects moving with varying foreground size in constant background. However, in ShanghaiTech Campus dataset, the normality representation is more complex as it contains varying foreground size with varying backgrounds.
\par \textbf{Ablation study:} 
To understand the effect of the features used and the proposed use of self-context in our approach, an ablation study is performed using the Avenue dataset. 
Table~\ref{ablation-table} lists the results, where it is observed that Model I ,which uses only the temporal features, performs better than Model II, which uses only the spatial features. Temporal features may be more important as anomalies in videos are mostly described by movements. Additionally, we observe that Model IV using both the feature types performs better than Model II, indicating the positive contribution of spatial features, which capture variations in the target /object appearance. Finally, consider the improvement achieved by Model IV (our final model) compared to Model III. The use of the input to the encoder into the decoder as well to form the self-context in our model is not considered in Model III, where a single pipeline with 2 layers of MSA \& LN and MLP \& LN modules are only used. The improvement highlights the significance of our proposed exploitation of a self-context for VAD.

%Model IV, which uses a transformer encoder-decoder performs better than Model III, which uses only transformer encoder as prediction module followed by a linear layer.   
\begin{table}[h]
	\begin{center}
	\vspace{-.2cm}
	\caption{\label{ablation-table} \small An ablation study of our approach on the CUHK Avenue dataset. DSF - Self-Context in the Decoder.}
	\vspace{-.3cm}
	\begin{tabular}{|m{1.4cm}|P{.9cm}|P{1.1cm}|P{.9cm}|P{1.1cm}|}
	\hline
\small \small{Model}& \small{ResNet}&\small{FlowNet}&\small{DSF}&\small{AUC $\%$}  \\  \hline
\small{Model I}&\small{$\times$}&\small{$\checkmark$}&\small{$\checkmark$}&\small{79.5}\\\hline
\small{Model II}&\small{$\checkmark$}&\small{$\times$}&\small{$\checkmark$}&\small{55.0}\\\hline
\small{Model III}&\small{$\checkmark$}&\small{$\checkmark$}&\small{$\times$}&\small{81.5}\\\hline
\small{Model IV}&\small{$\checkmark$}&\small{$\checkmark$} &\small{$\checkmark$}&\small{\textbf{89.4}}\\\hline
	\end{tabular}
	\end{center}
	 \vspace{-1.2cm}
\end{table}
%w/oSF - Prediction without self-context \hspace{.05cm} wSF - Prediction with self-context
\section{Conclusion}
\label{sec:conclusion}
A video anomaly detection approach has been proposed based on a one-class few-shot learning driven transformer prediction network that considers a self-context. Our learning strategy works on the video at hand, which not only reduces the training data requirement but also allows the capture of the video-relevant non-anomalous nature.
%We combined features extracted using pretrained ResNet and FlowNet2 to represent video frames. 
%We trained the transformer based prediction network using feature vectors of only few consecutive non-anomalous frames for each video thus eliminating the need of huge training dataset. 
%The predicted feature vector of a frame is compared with the actual feature vector of the frame to detect anomaly with imposed temporal consistency. 
Our approach has been found to perform well in comparison to the state-of-the-art on videos with different anomaly varieties, with the use of self-context resulting in a significant performance increase.

%Experimental results demonstrate the effectiveness of the proposed approach on videos. The qualitative results shows that a variety of anomalies in scenes with different backgrounds were detected successfully.
 %AUC value of 98.6$\%$ and 89.4$\%$ has been obtained for the UCSD Ped2 dataset and CUHK Avenue dataset respectively. AUC value of 80.6$\%$ have been obtained for ShanghaiTech Campus dataset.

\section{Aknowledgement}
Debashis Sen acknowledges the Science and Engineering Research Board (SERB), India for its assistance.

%\bibliographystyle{IEEE}
%\bibliography{main}

\end{document}